\documentclass[letterpaper, 10 pt, conference]{ieeeconf}  

\IEEEoverridecommandlockouts                              

\usepackage{graphics} 
\usepackage{epsfig} 
\usepackage{mathptmx} 
\usepackage{times} 
\usepackage{amsmath} 
\usepackage{amssymb}  
\usepackage[utf8]{inputenc}
\usepackage{booktabs, multirow, makecell}
\usepackage{cite}
\usepackage{siunitx}
\usepackage[table]{xcolor}
\usepackage{makecell}
\usepackage{bbold}

\usepackage[hidelinks]{hyperref}
\usepackage[capitalize]{cleveref}
\usepackage{censor}

\title{\LARGE \bf
Boundary-by-Mask: Few-Shot Instance Segmentation\linebreak[3] with Mask-Conditioned Boundary Learning \linebreak[3] for Texture-Poor Industrial Parts
}

\newif\ifanonymous
\anonymousfalse

\ifanonymous
\author{}
\else
\author{
Yutaka Yoshinaga$^{1}$ ,
Naoya Chiba$^{2}$ and
Koichi Hashimoto$^{3}$%
\thanks{$^{1}$Yutaka Yoshinaga is with the Graduate School of Information Sciences, Tohoku University, Sendai, Japan.
        {\tt\small yoshinaga.yutaka.t8@dc.tohoku.ac.jp}}%
\thanks{$^{2}$Naoya Chiba is with the Cybermedia Center, Osaka University, Osaka, Japan.
        {\tt\small chiba@nchiba.net}}%
\thanks{$^{3}$Koichi Hashimoto is with the Graduate School of Information Sciences, Tohoku University, Sendai, Japan.
        {\tt\small koichi.hashimoto.a8@tohoku.ac.jp}}}
\fi

\begin{document}
\maketitle
\thispagestyle{empty}
\pagestyle{empty}

\begin{abstract}
Recent advances in large pre-trained models have led to remarkable progress in instance segmentation on general images. However, industrial scenarios remain challenging. Instance definitions are often application-specific and inconsistent, and the domain gap from general imagery is substantial due to weak textures and limited contextual cues. Consequently, a direct application of existing models is unreliable. We propose Boundary-by-Mask, a few-shot instance segmentation framework that supervises boundaries instead of interior appearance. Given a few RGB images and corresponding instance masks, the method extracts rich visual features using a foundation-model encoder and trains a lightweight Signed Distance Function (SDF) head to predict boundary-aware distance maps. Segmentation masks are obtained through an SDF-to-mask reconstruction process. By explicitly estimating contours, the framework achieves reliable instance separation even on low-texture and color-uniform surfaces. The instance definition is conditioned by the instance mask. Replacing the mask specifies the segmentation target, such as the whole object or a sub-part.  A pixel-wise shallow MLP head enables rapid training. Experiments on industrial parts and food items with ambiguous boundaries show strong few-shot generalization, robustness in feature-poor conditions, and precise control over mask-level targets. Code and data will be released publicly.
\end{abstract}
\section{Introduction}
\label{sec:intro}

\begin{figure}[t]
  \centering
  \includegraphics[width=1.0\columnwidth]{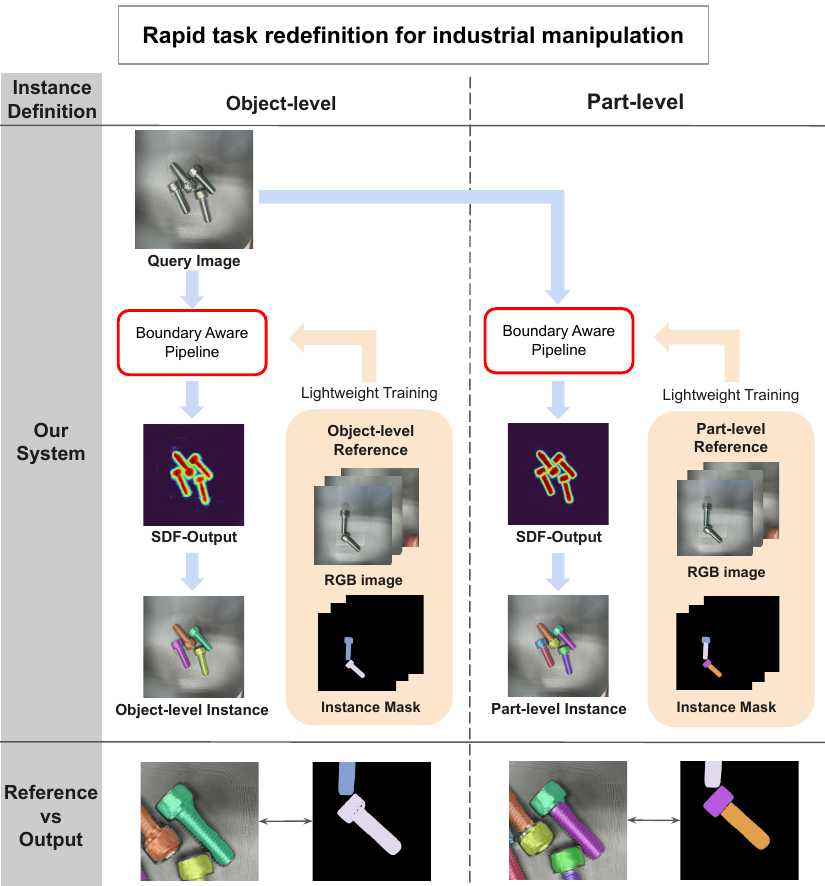}
  \caption{With a few RGB–mask references, our Boundary Aware Pipeline is lightly trained to predict a SDF on a query image and reconstruct instances. Simply changing the reference instance masks (object-level vs. part-level) switches the output granularity—no heavy retraining is required.}
  \label{fig:our-method1}
\end{figure}

Instance segmentation has become one of the central problems in computer vision, and remarkable progress has been achieved with the advent of large-scale datasets and foundation models such as Mask R-CNN \cite{he2018maskrcnn}, DETR \cite{carion2020endtoendobjectdetectiontransformers}, and the Segment Anything Model (SAM) \cite{kirillov2023segment}. These models have demonstrated impressive generalization to natural images. However, transferring their success to industrial domains remains challenging. The difficulty arises from both a visual gap and a semantic gap: industrial objects lack rich appearance cues, and the definition of an “instance” is application-dependent.

In industrial settings, the meaning of an instance changes according to its functional role.
For example, a component may need to be segmented into parts for an assembly task but treated as a whole in an inspection task.
Thus, the segmentation target itself depends on the task context. A segmentation model that can flexibly adapt to such varying instance definitions is essential for practical automation, where task requirements may change frequently.
In such environments, it is desirable that the model can be adapted rapidly to new instance definitions with only a few examples. 

Few-shot instance segmentation (FSIS) naturally supports such adaptability by learning from only a few reference examples.
Yet, most FSIS approaches rely on appearance-based feature matching between support and query images \cite{fan2020fgnfullyguidednetwork,zhang2023personalizesegmentmodelshot,mo2018partnetlargescalebenchmarkfinegrained,Wang_2019_ICCV}. This paradigm is effective for natural images with rich textures but becomes unreliable for industrial components, where appearance cues are weak and part boundaries are subtle or ambiguous. In such cases, the boundary rather than the interior appearance contains the most discriminative information \cite{Zitnick2014EdgeBL,Kervadec_2021}.

Distance-transform-based formulations have been explored for instance segmentation, such as the Deep Watershed Transform \cite{Bai_Urtasun_2017_CVPR_DeepWatershed}, where boundaries are inferred from learned energy maps. However, their application in few-shot, task-dependent industrial settings remains underexplored.

We therefore take a boundary-oriented perspective. Instead of predicting instance interiors, our goal is to predict \textit{where the boundary lies}. To represent boundaries continuously and robustly, we employ the signed distance function (SDF) as a supervisory signal. The SDF formulation encodes the boundary as the zero level set, enabling smooth interpolation and fine-grained control of object contours. 

Recent foundation encoders capture rich geometry and context, enabling boundary prediction from only a few annotated examples. We present \textbf{Boundary-by-Mask}, a few-shot instance segmentation framework that predicts a boundary-aware SDF conditioned on reference instance masks (Fig.~\ref{fig:our-method1}). A frozen encoder extracts dense features from a small set of RGB--mask pairs, and a pixel-wise MLP regresses the SDF map. By simply changing the reference mask, the target instance definition can be reconfigured without redesigning the model, realizing application-dependent segmentation. Training is lightweight (minutes) and parameter efficient, making the approach suitable for industrial changeovers. The boundary formulation further improves instance separation in texture-poor scenes.

In summary, our contributions are threefold:
\begin{itemize}
\item We formulate few-shot instance segmentation for industrial parts under frequent production changeovers, where instance definitions are application-dependent and boundary cues dominate over appearance cues.
\item We propose a boundary-aware framework that learns SDF from a few reference instance masks on top of a frozen foundation-model encoder, enabling rapid task-level redefinition without backbone retraining.
\item We demonstrate that the proposed method achieves robust low-shot performance, precise boundary reconstruction, and flexible instance specification simply by replacing the reference instance masks.
\end{itemize}
In addition, we constructed a few-shot instance segmentation dataset that accounts for cases where the definition of an instance varies across tasks.

\section{Related work}
\label{Related work}

\subsection{Few-Shot Instance Segmentation (FSIS)}

FSIS aims to segment unseen object categories from only a few annotated examples, referred to as \textit{support images}. 
Early methods extended Mask R-CNN with metric learning or feature matching between support and query images, such as Siamese Mask R-CNN~\cite{michaelis2019oneshotinstancesegmentation} and FGN~\cite{fan2020fgnfullyguidednetwork}. 
Subsequent works improved feature guidance and category adaptation~\cite{ganea2021incrementalfewshotinstancesegmentation,Wang2022DynamicTF}.

With the rise of large-scale foundation models, new few-shot and one-shot segmentation paradigms have emerged. 
Representative examples include Personalize Segment Anything~\cite{zhang2023personalizesegmentmodelshot}, SAM-IF~\cite{zhou2024samifleveragingsamincremental}, and MATCHER~\cite{liu2023matcher}, which leverage the promptable feature space of the SAM. 
No time to train!~\cite{espinosa2025timetraintrainingfreereferencebased} further demonstrates a training-free reference-matching approach.  

Despite architectural differences, most FSIS methods still rely on appearance-based feature matching between support and query. 
This paradigm becomes unreliable for \textbf{industrial parts}, which exhibit texture-poor, color-uniform, and application-dependent semantics. 
Such visual uniformity and semantic variability limit the robustness of conventional FSIS, motivating the exploration of \textbf{boundary-aware} alternatives discussed in the following sections.

\begin{figure*}[t]
  \centering
  \includegraphics[width=\textwidth]{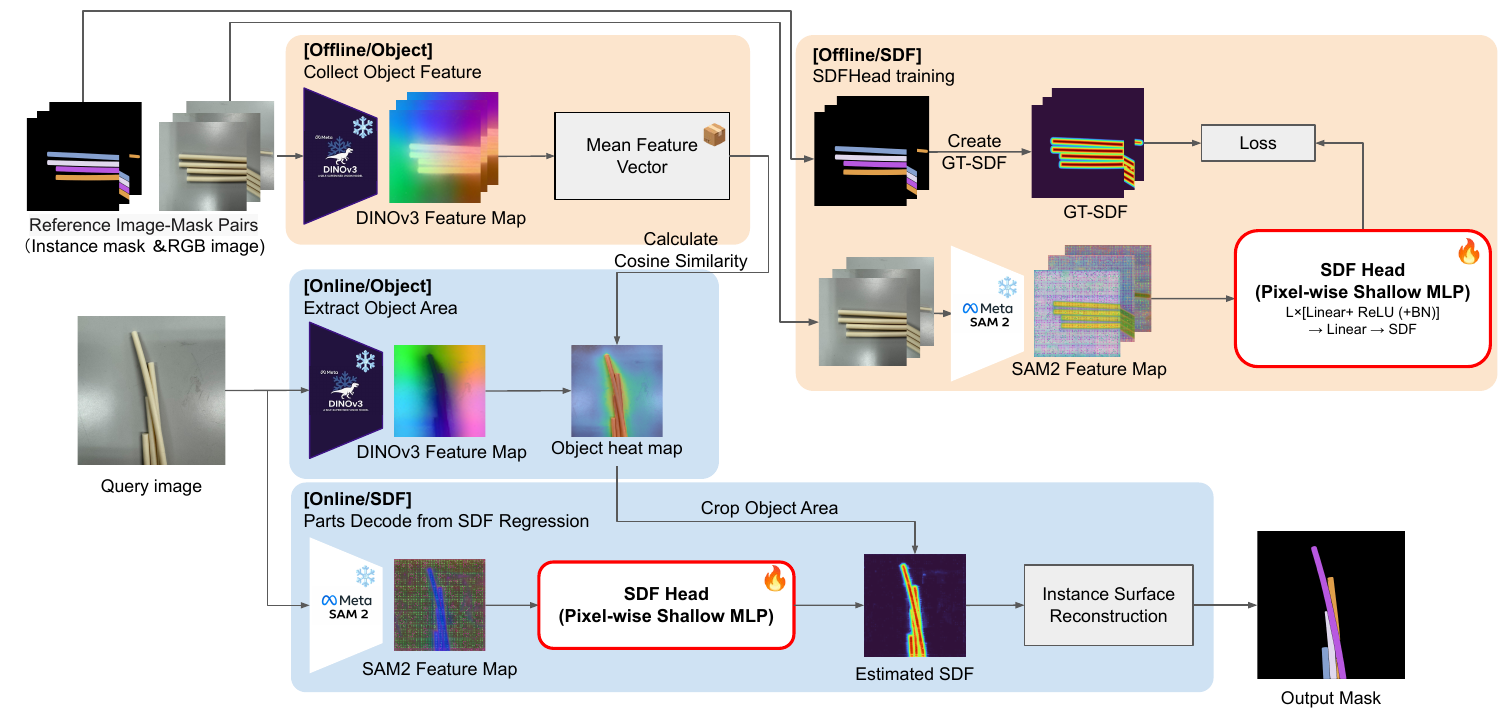}
  \caption{Overview of the method. The inputs are a few RGB–mask reference pairs and a query RGB image; the output is the instance mask for the query. The pipeline consists of four blocks (Offline/Online × Object/SDF): Offline/Object builds an object feature vector with DINOv3; Offline/SDF converts reference masks to GT-SDF and trains a pixel-wise shallow MLP on SAM2 features. Online/Object localizes the query object area, and Online/SDF regresses an SDF within that area and reconstructs the final instance masks.}
  \label{fig:our-method}
\end{figure*}

\subsection{Vision Foundation Models for Segmentation}
The rapid progress of vision foundation models has reshaped how segmentation tasks are approached.
Self-supervised vision transformers such as DINO~\cite{caron2021emergingpropertiesselfsupervisedvision,dinov2,dinov3} and masked image modeling frameworks like MAE~\cite{he2021maskedautoencodersscalablevision} learn object-centric and context-rich visual representations without labels, capturing structural regularities and inter-part relationships useful for dense prediction.
In parallel, segmentation-oriented foundation models including Mask2Former~\cite{cheng2022maskedattentionmasktransformeruniversal}, SEEM~\cite{zou2023segment}, and the Segment Anything family~\cite{kirillov2023segment,ravi2024sam2segmentimages,carion2025sam3segmentconcepts} focus on geometric precision and boundary localization through promptable interfaces.

Recent studies have explored their complementary use.  
No time to train!~\cite{espinosa2025timetraintrainingfreereferencebased} employed SAM to segment candidate regions and matched each mask with DINO features for training-free instance segmentation, revealing the strong complementarity between DINO’s contextual encoding and SAM’s geometric segmentation.  
Building on this insight, recent works increasingly leverage DINO-like encoders for contextual understanding and SAM-like models for precise boundary localization~\cite{ren2024groundedsamassemblingopenworld}, forming hybrid representations that balance structure and geometry—particularly suitable for texture-poor industrial parts.

\subsection{Industrial Semantics and Application-Dependent Segmentation}

Unlike everyday objects, industrial components lack fixed semantic categories.  
Their meaning varies with the operational context—assembly, manipulation, or inspection.  
For example, a component may need to be segmented into parts for an assembly task but treated as a whole in an inspection task.
Such variability demands segmentation that adapts to task context rather than relying on static class labels.

Recent studies have explored this semantic flexibility through context- or application-dependent segmentation.  
Spider~\cite{zhao2024spiderunifiedframeworkcontextdependent} addressed context-dependent concept segmentation, and part-aware or hierarchical approaches~\cite{mo2018partnetlargescalebenchmarkfinegrained,feng2022robustpartawareinstancesegmentation,zhu2025hierarchical} emphasized adapting segmentation granularity to task needs in industrial scenes.  
However, most remain rule-based or geometry-specific, lacking dynamic adaptability to unseen contexts.

We extend this notion to few-shot instance segmentation, formulating segmentation as an application-dependent, reference-driven process.  
By leveraging foundation-model features and boundary-aware learning, our framework aims to achieve flexible, task-aligned segmentation for industrial manipulation and inspection.

\section{Methodology}
In this section, we formalize the problem and present an overview of \textbf{Boundary-by-Mask}.
The method comprises four complementary segments.
We first outline the overall architecture, and then describe each segment’s role and processing in turn,
thereby clarifying the design rationale and operating principles of the proposed approach.

\subsection{Method Overview}
We propose \textbf{Boundary-by-Mask}, a few-shot framework for \textbf{arbitrary instance segmentation of industrial parts} that learns \textbf{boundary-aware SDF} on top of a foundation-model encoder. 
The goal is to segment both entire objects and their sub-parts reliably in \emph{feature-sparse} conditions such as low texture or color-uniform. 
The model is trained using a small set of RGB images and instance masks. During inference, it estimates instance masks for RGB images.

The method operates in two phases: an offline phase for feature preparation and SDF training, and an online phase for query-time inference.
In the \emph{offline phase}, we use DINOv3\cite{dinov3} to extract features from reference images and compute a \emph{mean feature vector} for object localization.
In parallel, the reference instance masks are converted into ground-truth SDF (GT-SDF), and a lightweight \emph{SDF head}—a pixel-wise shallow MLP built on top of the SAM2\cite{ravi2024sam2segmentimages} encoder—is trained to predict per-pixel SDF values.
Because the model focuses on boundaries instead of interior texture, it becomes more robust to weak appearance cues.
In the \emph{online phase}, a query image is first localized by matching its DINOv3 features to the  reference mean feature vector.
Within the localized region, the trained SDF head predicts the SDF, from which instance masks are reconstructed through \textbf{Instance Surface Reconstruction}.

Each phase contains two complementary branches, forming four modules in total. 
The \emph{Object branch} extracts and localizes object regions, while the \emph{SDF branch} learns and applies the boundary-aware representation for instance mask decoding. 
The segmentation granularity is \emph{mask-conditioned}: by changing the reference instance mask, the system can switch between object-level and part-level segmentation simply by retraining the lightweight SDF head. 
The following sections describe each component in detail; an overview of the pipeline is shown in Fig.~\ref{fig:our-method}.

\subsection{Collect Object Feature}
The \textbf{Collect Object Feature} module corresponds to the \textbf{Offline / Object branch}. 
It builds a set of reference features that later enable reliable object localization during online inference. 
By pre-extracting representative object features from reference images using DINOv3, the model can localize objects robustly even under weak texture conditions.

Let $\mathbf{I}_r$ be a reference image and $\mathbf{M}_r^{(p)}$ its binary mask for part $p$.
The DINO feature map of the reference image is extracted by the DINOv3 encoder.
We then compute the mean feature vector $\mathbf{f}_r^{(p)}$ within the mask region by element-wise multiplication.
This averaging reduces local noise and produces a compact, texture-invariant descriptor.
Descriptors are then $L_2$-normalized and stored as the \textbf{mean feature vector}.

\subsection{SDF Head Training}
The \textbf{SDF Head Training} module corresponds to the \textbf{Offline / SDF branch}, 
which learns a \textbf{boundary-aware representation} capable of reconstructing instance surfaces of texture-poor industrial parts. 
Unlike conventional segmentation models that focus on appearance, this branch supervises object \emph{boundaries} using SDF.

\begin{figure}
  \makebox[\columnwidth][c]{%
    \includegraphics[width=0.95\columnwidth]{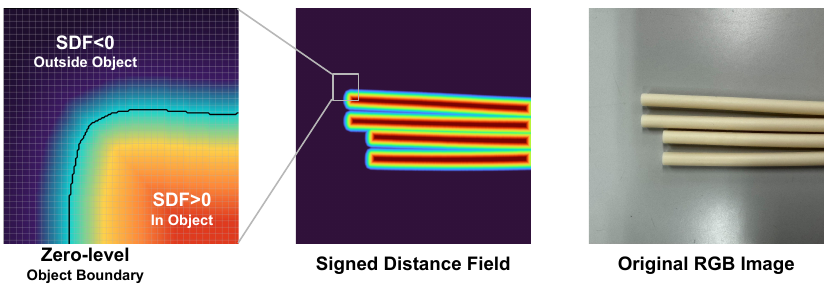}%
  }
\caption[SDF around tubes and the source RGB image.]{
\textbf{SDF convention and example.}
\textbf{Left:} schematic—object boundary is the zero level; $\mathrm{SDF}>0$ inside and $\mathrm{SDF}<0$ outside.
\textbf{Middle:} SDF of the tube objects (normalized signed distance; the zero level aligns with tube boundaries).
\textbf{Right:} the original RGB image from which the SDF was computed.
}
  \label{fig:-SDF_Definition}
\end{figure}

\subsubsection*{Ground-truth SDF Generation}
Given a binary mask $\mathbf{M}_r$, we first generate an SDF $\mathbf{D}_r$ that takes zero on the object boundary, has magnitude equal to the Euclidean distance from the boundary, and is positive inside the object while negative outside~\cite{1467575,DANIELSSON1980227} (Fig.~\ref{fig:-SDF_Definition}).

To suppress large magnitude far from boundaries, we use a truncated and normalized signed distance field.
Given a truncation radius $\tau>0$, we clip $\mathbf{D}_r$ to $[-\tau,\tau]$ and normalize it to $[-1,1]$, yielding $\tilde{\mathbf{D}}_r$, which is used as the training target.

\subsubsection*{Network Architecture}
The SDF predictor consists of a frozen \textbf{SAM2} encoder and a lightweight pixel-wise MLP head.
We extract the \textbf{SAM2} feature map and apply a shared pixel-wise MLP (Linear--ReLU--BN) independently at each spatial location to regress the SDF, enabling fast mask-conditioned adaptation.

\subsubsection*{SDF Loss}
We compute the SDF loss between the clipped SDF head output
$\tilde{\mathbf{D}}_{r}^{\ast} \in [-1,1]$
and the ground-truth truncated and normalized map $\tilde{\mathbf{D}}_r$
using a Smooth-$\ell_1$ loss:
\[
\mathcal{L}_{\text{SDF}}
= \mathrm{SmoothL1}\!\left(
\tilde{\mathbf{D}}_{r}^{\ast},\,
\tilde{\mathbf{D}}_r
\right).
\]
Restricting the prediction range to $[-1,1]$ emphasizes the boundary region
and prevents large interior distances from dominating the gradients.
This boundary-centered supervision stabilizes training
and improves generalization to texture-poor objects.

\subsubsection*{Threshold Estimation for SDF Seed Cut}
To convert the continuous SDF prediction into discrete instance seeds,
a threshold $\tau_{\text{obj}}$ is required.
This threshold determines how deep inside the object
the seed region is defined and directly affects
instance separation quality.

To determine the optimal threshold $\tau_{\text{obj}}$, we perform a greedy search on the reference set.

For each reference pair $(\mathbf{I}_r,\mathbf{M}_r)$, the trained SDF head outputs $\tilde{\mathbf{D}}_{r}^{\ast}$, 
which is binarized with various thresholds $\tau$ as
$
\mathbf{B}_r^{(\tau)} = \mathbb{1}[\tilde{\mathbf{D}}_{r}^{\ast} > \tau].
$
Each candidate mask is evaluated against the ground truth using a composite score:
\begin{equation}
\label{tau}
\begin{aligned}
S(\tau;\,\tilde{\mathbf{D}}_{r}^{\ast},\mathbf{M}_r)
  &= w_c S_c(\mathbf{B}_r^{(\tau)},\mathbf{M}_r)
   + w_b S_b(\mathbf{B}_r^{(\tau)},\mathbf{M}_r)\\
  &\quad + w_u S_u(\mathbf{B}_r^{(\tau)},\mathbf{M}_r)
   - w_f S_f(\mathbf{B}_r^{(\tau)},\mathbf{M}_r).
\end{aligned}
\end{equation}

where $S_c$ measures instance count consistency to ensure the number of predicted objects matches the ground truth, 
$S_b$ evaluates boundary alignment using the boundary F1 score, 
$S_u$ measures the overall overlap between prediction and ground truth via the union IoU, 
and $S_f$ penalizes excessive fragmentation of a single object into multiple small pieces.

The optimal threshold is then chosen as
\[
\tau_{\text{obj}}^{\ast} = \arg\max_{\tau} S(\tau),
\]
which aligns the SDF cut with the true object boundaries. 
This automatic estimation ensures robust and adaptive mask separation across diverse part geometries.

\begin{figure*}[t]
  \makebox[\textwidth][c]{%
    \includegraphics[width=0.95\textwidth]{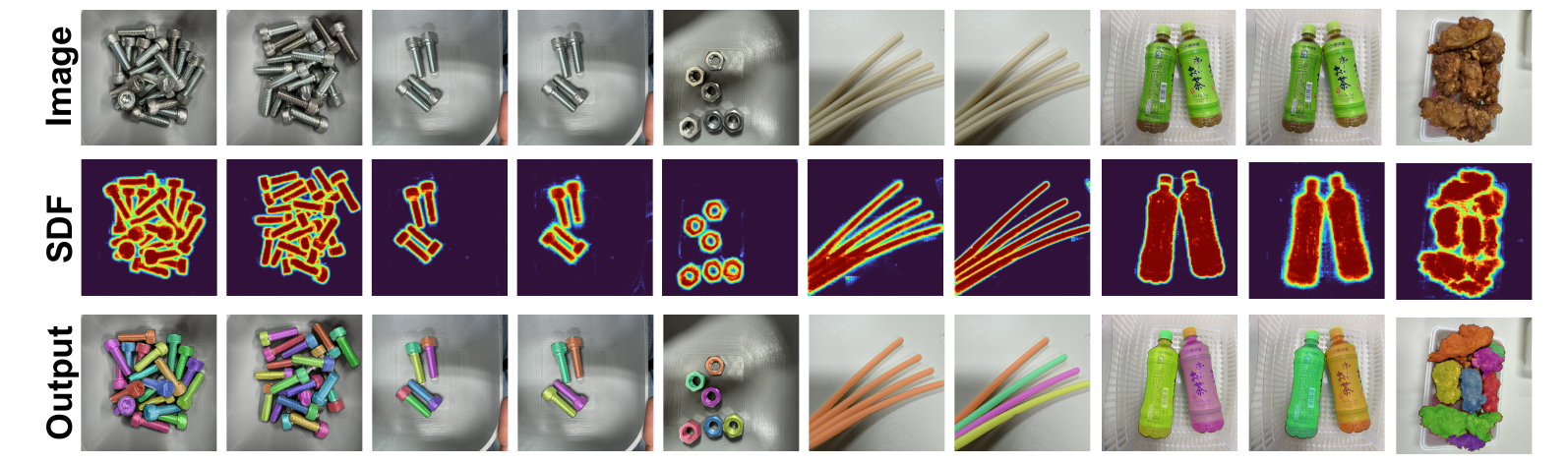}%
  }
\caption{Qualitative results of \textbf{Boundary-by-Mask}.
Top : query RGB, middle : predicted SDF, bottom : output masks overlaid on the query.
}
  \label{fig:our-result}
\end{figure*}

\begin{figure*}[t]
  \makebox[\textwidth][c]{%
    \includegraphics[width=0.87\textwidth]{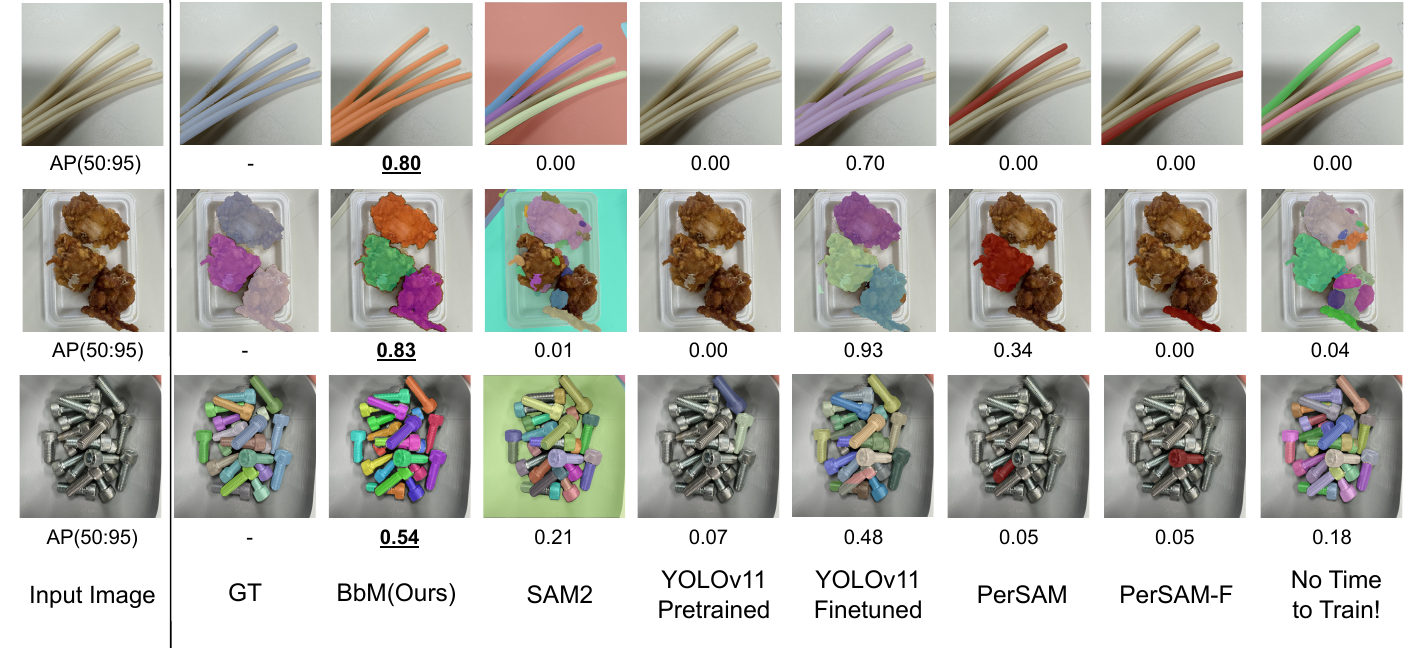}%
  }
\caption[Qualitative comparison on three queries; AP is computed only on these images.]{
\textbf{Qualitative comparison on three queries} (top: tubes, middle: fried chicken, bottom: screw pile).
Left to right: input image, ground truth (GT), \textbf{Boundary-by-Mask} (BbM; ours),
and baselines—SAM2 (AMG), YOLOv11 (pretrained), YOLOv11 (finetuned),
PerSAM (no fine-tuning), PerSAM-F, and No time to train!.
\emph{Shown AP values are per-image scores, computed independently for each image.}
}
  \label{fig:-compare}
\end{figure*}

\begin{figure}[t]
  \centering
  \includegraphics[width=0.7\columnwidth]{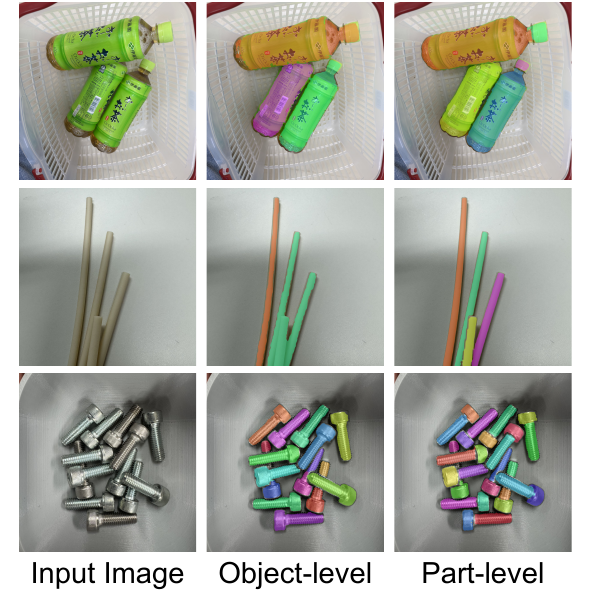}
\caption{\textbf{Mask-conditioned flexibility.} By changing only the training targets for the SDF Head, the predicted instances adapt accordingly. Left to right: input image, object-level segmentation, part-level segmentation.}
  \label{fig:object-part}
\end{figure}

\subsection{Extract Object Area}
The \textbf{Extract Object Area} module corresponds to the \textbf{Online / Object branch}, 
which localizes object regions in a query image by matching features with the reference mean feature vector.

A query image $\mathbf{I}_q$ is encoded by DINOv3 to obtain dense feature map $\mathbf{F}_{\text{DINO}}^{q}$.
It is compared with reference mean feature vector $\mathbf{f}_r^{(p)}$ 
using cosine similarity.
The resulting similarity map is smoothed and thresholded at $\tau_{\text{loc}}$ 
to produce a localized region $\mathbf{R}_q=\{(x,y)\mid S(x,y)>\tau_{\text{loc}}\}$.
Small isolated regions are removed to maintain spatial coherence. 
This simple feature-matching step provides a reliable crop 
for the subsequent SDF-based mask reconstruction.

\subsection{Surface Reconstruction from SDF Regression}
The \textbf{Surface Reconstruction from SDF Regression} module corresponds to the \textbf{Online / SDF branch}, 
which reconstructs instance masks from the boundary-aware SDF predicted by the SDF head. 
This process, called \textbf{Instance Surface Reconstruction}, 
converts continuous SDF predictions into discrete object masks.

Given a query image $\mathbf{I}_q$, 
the localized region $\mathbf{R}_q$ is encoded by SAM2 to obtain dense embeddings $\mathbf{F}_{\text{SAM2}}^q$, 
which are passed to the trained SDF Head to predict a signed distance map
$
\tilde{\mathbf{D}}_{q}^{\ast} = \mathcal{H}_{\text{SDF}}(\mathbf{F}_{\text{SAM2}}^q).
$
The map $\tilde{\mathbf{D}}_{q}^{\ast}$ takes values in $[-1,1]$, 
where the zero-level contour represents the estimated object boundary.

\subsubsection*{Seed Extraction and Adaptive Refinement}
To extract instances, the predicted SDF is thresholded as seed points
$
\Omega_{\text{seed}} = \{(x,y)\ |\ \tilde{\mathbf{D}}_{q}^{\ast}(x,y) > \tau_{\text{obj}}\}
$
with the learned cut value $\tau_{\text{obj}}$ obtained from the reference phase.
Each connected component above a minimum area is treated as an instance seed. 
These seeds serve as initialization for an SDF-guided refinement process that reconstructs complete and non-overlapping object masks.

An energy map combines the inverted SDF with image-gradient cues:
\[
E(x,y) = -\tilde{\mathbf{D}}_{q}^{\ast}(x,y) + \lambda\,\|\nabla I_q(x,y)\|,
\]
where the first term pulls pixels toward SDF basins (object interiors) and the second discourages crossing strong edges.
We then apply a watershed transform\cite{Vincent1991WatershedsID} with seed markers $\Omega_{\text{seed}}$:
\[
\mathbf{L}_q=\mathrm{Watershed}\bigl(E,\ \text{markers}=\Omega_{\text{seed}}\bigr),
\]
$\mathbf{L}_q$ is the instance label map (0: background; $1,2,\dots$: instance IDs),
producing instance-consistent regions bounded by predicted SDF contours.

\begin{table}[t]
  \centering
  \caption{Dataset composition (test split): category, layout, granularity, \#images, and Inst./img (mean$\pm$std).}
  \label{tab:dataset-composition}
  \setlength{\tabcolsep}{6pt}
  \sisetup{detect-weight=true,detect-inline-weight=math}
  \footnotesize
  \resizebox{\columnwidth}{!}{%
  \begin{tabular}{l l l S[table-format=3.0] c}
    \toprule
    \textbf{Category} & \textbf{Layout} & \textbf{Granularity} &
    \textbf{Test imgs} & \textbf{Inst./img (mean $\pm$ std)} \\
    \midrule
    Screws        & flat & object-level & 45 & \(4.29\pm0.58\) \\
    Screws        & flat & part-level   & 42 & \(7.95\pm1.66\) \\
    Screws        & pile & object-level & 55 & \(16.40\pm3.91\) \\
    Screws        & pile & part-level   & 40 & \(30.12\pm6.83\) \\
    PET bottles   & flat & part-level   & 41 & \(4.88\pm1.44\) \\
    PET bottles   & flat & object-level & 37 & \(2.41\pm0.79\) \\
    Tubes         & flat & object-level & 35 & \(1.23\pm0.59\) \\
    Tubes         & flat & part-level   & 35 & \(3.69\pm1.03\) \\
    Hex nuts      & flat & object-level & 40 & \(4.10\pm1.07\) \\
    Fried chicken & flat & object-level & 36 & \(2.44\pm0.64\) \\
    Fried chicken & pile & object-level & 34 & \(5.44\pm1.14\) \\
    \midrule
    \textit{flat (subtotal)} & flat &  & 311 & \(3.99\pm2.19\) \\
    \textit{pile (subtotal)} & pile &  & 129 & \(17.77\pm10.47\) \\
    \textbf{Total (test)}    &      &  & 440 & \(8.03\pm8.65\) \\
    \bottomrule
  \end{tabular}}
\end{table}

\begin{table}[t]
  \centering
  \begingroup
  \setlength{\tabcolsep}{3pt}
  \renewcommand{\arraystretch}{0.90}
  \setlength{\aboverulesep}{0.3ex}
  \setlength{\belowrulesep}{0.3ex}
  \caption{We compare BbM against several baselines. BbM remains highly competitive across all $K$ and scales well with more references; at $K{=}10$ it achieves \textbf{AP(50:95)=0.53} with strong \textbf{AP75=0.66} and \textbf{mIoU=0.92}, reflecting precise boundaries and clean region overlap.}
  \label{tab:bbm-compare-selected}
  \scriptsize
  \begin{tabular}{l S[table-format=1.2] S[table-format=1.2] S[table-format=1.2] S[table-format=1.2]}
    \toprule
    \textbf{Method (setting)} &
    \multicolumn{1}{c}{\makecell{\textbf{COCO}\\\textbf{AP(50:95)}}} &
    \multicolumn{1}{c}{\textbf{AP50}} &
    \multicolumn{1}{c}{\textbf{AP75}} &
    \multicolumn{1}{c}{\textbf{mIoU}} \\
    \midrule

    \rowcolor{gray!15}\multicolumn{5}{l}{\bfseries Zero-shot}\\[-0.8ex]
    \cmidrule(lr){1-5}
    SAM2 \cite{ravi2024sam2segmentimages} (large)              & 0.18 & 0.28 & 0.18 & 0.16 \\
    YOLOv11-seg\cite{khanam2024yolov11overviewkeyarchitectural} (pretrained)  & 0.15 & 0.21 & 0.16 & 0.34 \\
    \midrule[0.9pt]

    \rowcolor{gray!15}\multicolumn{5}{l}{\bfseries K = 1}\\[-0.8ex]
    \cmidrule(lr){1-5}
    PerSAM \cite{zhang2023personalizesegmentmodelshot} (no-FT)            & 0.11 & 0.15 & 0.11 & 0.59 \\
    PerSAM-F\cite{zhang2023personalizesegmentmodelshot}                 & 0.12 & 0.17 & 0.12 & 0.56 \\
    No time to train! \cite{espinosa2025timetraintrainingfreereferencebased}   & 0.13 & 0.29 & 0.01 & 0.89 \\
    YOLOv11-seg (m)\cite{khanam2024yolov11overviewkeyarchitectural}           & 0.20 & 0.38 & 0.16 & 0.69 \\
    \rowcolor{gray!05}\textbf{BbM (ours)} & \textbf{0.30} & \textbf{0.52} & \textbf{0.34} & \textbf{0.90} \\
    \midrule[0.9pt]

    \rowcolor{gray!15}\multicolumn{5}{l}{\bfseries K = 2}\\[-0.8ex]
    \cmidrule(lr){1-5}
    PerSAM \cite{zhang2023personalizesegmentmodelshot} (no-FT)            & 0.11 & 0.16 & 0.11 & 0.59 \\
    PerSAM-F\cite{zhang2023personalizesegmentmodelshot}                 & 0.12 & 0.16 & 0.11 & 0.56 \\
    No time to train!\cite{espinosa2025timetraintrainingfreereferencebased}      & 0.15 & 0.31 & 0.12 & 0.86 \\
    YOLOv11-seg (s)\cite{khanam2024yolov11overviewkeyarchitectural}           & 0.31 & 0.53 & 0.29 & 0.78 \\
    \rowcolor{gray!05}\textbf{BbM (ours)} & \textbf{0.43} & \textbf{0.68} & \textbf{0.52} & \textbf{0.91} \\
    \midrule[0.9pt]

    \rowcolor{gray!15}\multicolumn{5}{l}{\bfseries K = 5}\\[-0.8ex]
    \cmidrule(lr){1-5}
    PerSAM \cite{zhang2023personalizesegmentmodelshot} (no-FT)            & 0.10 & 0.15 & 0.10 & 0.59 \\
    PerSAM-F\cite{zhang2023personalizesegmentmodelshot}                 & 0.12 & 0.17 & 0.12 & 0.56 \\
       No time to train!\cite{espinosa2025timetraintrainingfreereferencebased}      & 0.15 & 0.31 & 0.11 & 0.82 \\
    YOLOv11-seg (s)\cite{khanam2024yolov11overviewkeyarchitectural}           & 0.43 & 0.66 & 0.44 & 0.81 \\
    \rowcolor{gray!05}\textbf{BbM (ours)} & \textbf{0.46} & \textbf{0.70} & \textbf{0.56} & \textbf{0.92} \\
    \midrule[0.9pt]

    \rowcolor{gray!15}\multicolumn{5}{l}{\bfseries K = 10}\\[-0.8ex]
    \cmidrule(lr){1-5}
    PerSAM \cite{zhang2023personalizesegmentmodelshot} (no-FT)            & 0.11 & 0.15 & 0.10 & 0.59 \\
    PerSAM-F\cite{zhang2023personalizesegmentmodelshot}                 & 0.12 & 0.16 & 0.12 & 0.56 \\
       No time to train!\cite{espinosa2025timetraintrainingfreereferencebased}      & 0.14 & 0.28 & 0.11 & 0.84 \\
    YOLOv11-seg (s)\cite{khanam2024yolov11overviewkeyarchitectural}           & \textbf{0.55} & 0.78 & 0.59 & 0.86 \\
    \rowcolor{gray!05}\textbf{BbM (ours)} & 0.53 & \textbf{0.80} & \textbf{0.66} & \textbf{0.92} \\
    \bottomrule
  \end{tabular}
  \endgroup
\end{table}

\section{Experiments}
\label{sec:experiments}

\subsection{Objective}

The objective of our experiments is to quantitatively verify the effectiveness and practical viability of the proposed Boundary-by-Mask (BbM) framework.
We focus on three aspects:
(i) how BbM compares with strong appearance-driven baselines,
(ii) how segmentation accuracy varies with the number of reference images,
and (iii) how boundary-aware SDF learning improves robustness and practical adaptability, as analyzed through component comparisons, parameter sensitivity, and training-time scalability.
Together, these analyses clarify the benefits of boundary-aware, mask-conditioned learning for few-shot instance segmentation of industrial parts.

\subsection{Experimental Setup}
\subsubsection*{Datasets and Evaluation Metric}
We evaluated BbM on a custom dataset comprising eleven subsets from five categories:
nut, screw, tube, fried chicken, and plastic bottle, with approximately 50 RGB--mask pairs per subset.

The categories are selected to reflect diverse geometric and material properties relevant to industrial automation.
Nuts represent non-convex rigid objects,
screws, bottles, and tubes include both object- and part-level definitions,
tubes model deformable and texture-poor components,
and fried chicken introduces irregular shapes with self-occlusion as a challenging test case beyond rigid parts.

The dataset further covers flat and pile layouts with varying instance densities.
Table~\ref{tab:dataset-composition} summarizes the dataset statistics.

Existing datasets such as PASCAL-Part~\cite{Chen_2014_CVPR} and PACO-Part~\cite{Ramanathan_2023_CVPR} provide part-level annotations in natural images,
while industrial datasets such as Siléane~\cite{Bregier_2017_ICCV} mainly focus on object-level segmentation.
Our dataset combines task-dependent granularity with layout variation,
enabling evaluation of application-dependent instance segmentation.

Performance is assessed using mIoU and COCO-style $\mathrm{AP}_{50{:}95}$, $\mathrm{AP}_{50}$, and $\mathrm{AP}_{75}$~\cite{lin2015microsoftcococommonobjects}.

\subsubsection*{Evaluation Setting}
\begin{figure}[t]
  \centering
  \includegraphics[width=0.95\columnwidth]{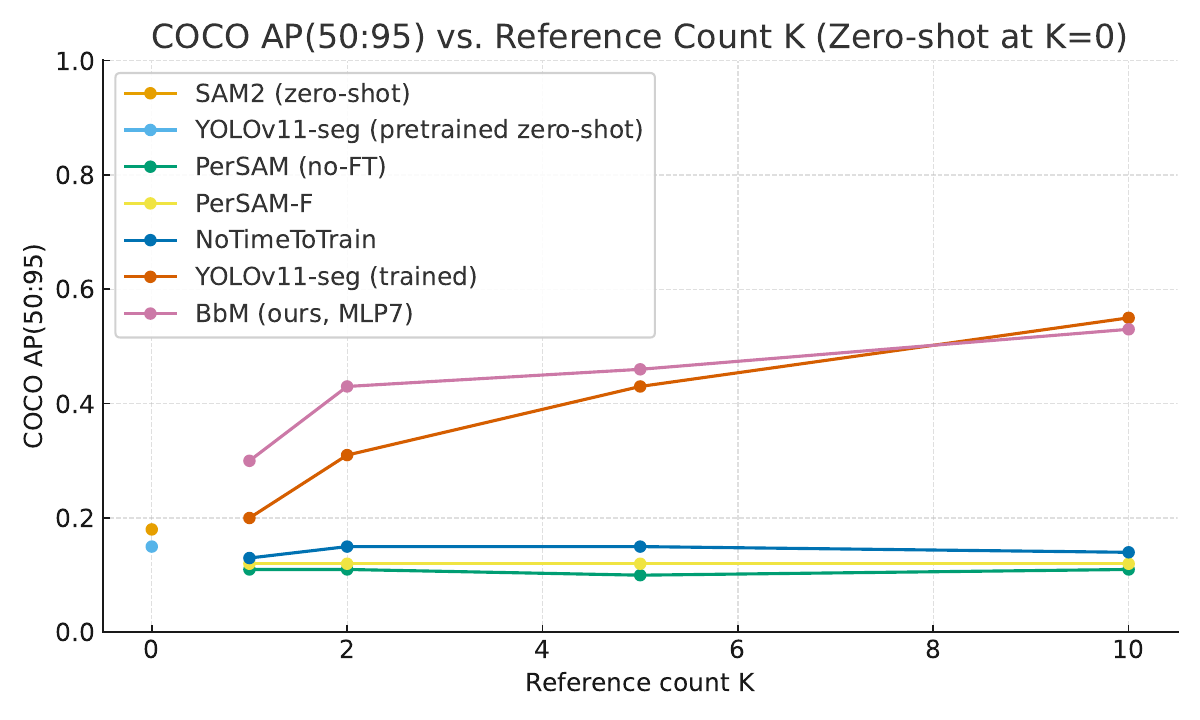}
\caption{\textbf{Accuracy vs.\ reference count $K$ (COCO $\mathrm{AP}_{50{:}95}$).} 
On our industrial-parts dataset, \textbf{BbM} (pink) rises steeply from $K{=}1$ to $K{=}5$ and then nearly saturates; it outperforms YOLOv11 for $K\le5$, while YOLOv11 slightly overtakes at $K{=}10$. }
  \label{fig:accuracy}
\end{figure}

We conduct three main evaluations aligned with the above objectives:
\begin{itemize}
  \item \textbf{(i) Cross-method comparison:} BbM vs.\ baselines (SAM2, YOLOv11~\cite{khanam2024yolov11overviewkeyarchitectural}, PerSAM, PerSAM-F, No time to train!) under matched splits and the same $K$ for fair FSIS comparison (Table~\ref{tab:bbm-compare-selected}, Fig.~\ref{fig:-compare}).
  \item \textbf{(ii) Reference-shot analysis:} accuracy trends as the number of reference RGB--mask pairs varies, $K \in \{1,2,5,10\}$ (Fig.~\ref{fig:accuracy}).
  \item \textbf{(iii) Robustness and practicality analyses:} (a) component comparison of SDF vs.\ direct mask/edge prediction, (b) sensitivity to key inference parameters, and (c) training-time scalability as a function of $K$ (Table~\ref{tab:head-ablation}, Table~\ref{tab:ablation-tau}, Fig.~\ref{fig:runtime-offline}).
\end{itemize}

All models are implemented in PyTorch and evaluated on a single NVIDIA RTX~A4000, using identical query splits and unseen query images for fairness.
Unless otherwise noted, we use the 7-layer MLP head (MLP7) as the default SDF head, trained with AdamW~\cite{loshchilov2018decoupled} and a learning rate of $3\times10^{-4}$.
The SDF seed threshold $\tau_{\text{obj}}$ is selected on the reference images by maximizing Eq.~(\ref{tau}) with $(w_c, w_b, w_u, w_{\text{f}}) = (1.0, 0.6, 0.3, 0.8)$.
Additional analyses for alternative settings are provided in the robustness experiments.

\subsection{Experimental Results}
We first summarize the main findings for (i)–(iii), then point to figures/tables as supporting evidence. Representative BbM outputs are in Fig.~\ref{fig:our-result}, and object/part flexibility is illustrated in Fig.~\ref{fig:object-part}.

\paragraph{(i) Cross-method comparison.}
BbM achieves consistently higher $\mathrm{AP}_{50}$, $\mathrm{AP}_{75}$, and mIoU than appearance-driven baselines, with the largest margins at small $K$ and on texture-poor parts; for $\mathrm{AP}_{50{:}95}$, YOLOv11 is slightly higher only at $K{=}10$, while BbM leads otherwise.
These results indicate that emphasizing boundaries is particularly beneficial under scarce reference supervision and weak interior cues (quantitative: Table~\ref{tab:bbm-compare-selected}, Fig.~\ref{fig:accuracy}; qualitative: Fig.~\ref{fig:-compare}). On the other hand, when a relatively large number of samples are available even in the few-shot setting ($K{=}10$), finetuned YOLOv11, which optimizes the entire network including the backbone, outperforms our method.
This suggests a turning point where scene-specific features become more beneficial than those provided by large pre-trained models.

\paragraph{(ii) Reference-shot analysis.}
Performance improves sharply from $K{=}1$ to $K{=}2$--$5$ and then saturates beyond $K{=}5$,
indicating that BbM captures boundary structure with a modest number of references (Fig.~\ref{fig:accuracy}).

\begin{figure}[t]
  \centering
  \includegraphics[width=0.70\columnwidth]{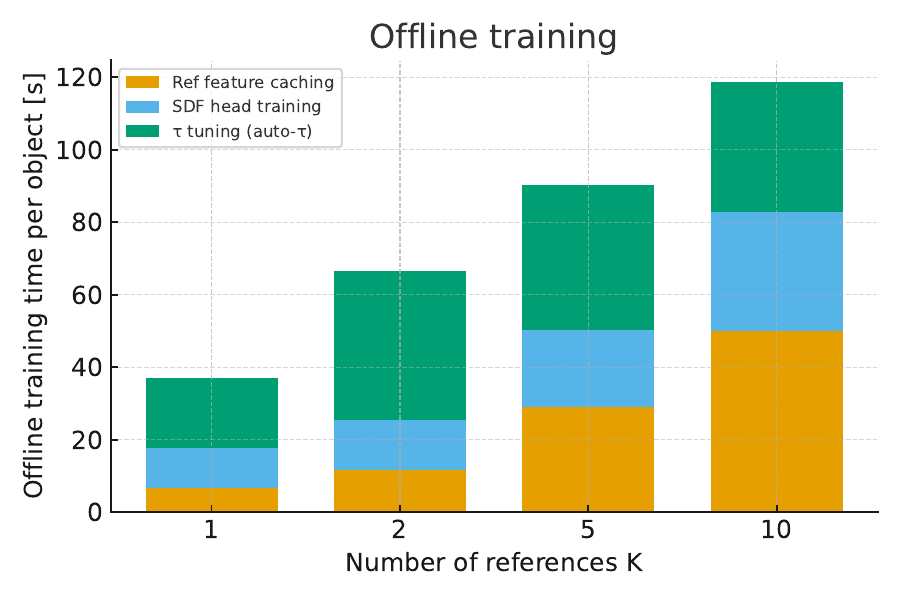}
  \caption{Offline reference-phase time for BbM as a function of $K$:
  feature caching, head training, and $\tau_{\text{obj}}$ auto-tuning on a single RTX~A4000.}
  \label{fig:runtime-offline}
\end{figure}

\paragraph{(iii) Robustness and practicality analyses.}
Table~\ref{tab:head-ablation} reports an apples-to-apples head ablation with identical frozen features and head capacity, where only the supervision target differs.
While direct mask or edge prediction yields limited gains, SDF supervision consistently provides large improvements across all $K$, confirming that boundary distance learning is central to BbM.

We also test the stability of the automatic $\tau_{\text{obj}}$ estimation by varying the scoring weights in Eq.~(\ref{tau}).
Table~\ref{tab:ablation-tau} shows that the results remain virtually unchanged across weight strategies, indicating low sensitivity of the threshold selection.
Finally, Fig.~\ref{fig:runtime-offline} shows that offline preparation remains small and scales mildly with $K$, supporting rapid task-level redefinition.
We measured the manual cost to annotate a reference mask as $\sim$15.4\,s per image.
Combining annotation with the offline reference phase, end-to-end adaptation completes within $\sim$54\,s for $K{=}1$, enabling efficient deployment.

\begin{table}[t]\small
\centering
\caption{Head ablation with identical frozen features and head capacity; only the supervision target differs.}
\vspace{-0.3em}
\label{tab:head-ablation}
\setlength{\tabcolsep}{1.6pt}
\begin{tabular}{lcc|cc|cc|cc}
\toprule
\multirow{2}{*}{Target} &
\multicolumn{2}{c|}{$K{=}1$} &
\multicolumn{2}{c|}{$K{=}2$} &
\multicolumn{2}{c|}{$K{=}5$} &
\multicolumn{2}{c}{$K{=}10$} \\
 & AP & mIoU & AP & mIoU & AP & mIoU & AP & mIoU \\
\midrule
Mask      & 0.03 & 0.38 & 0.03 & 0.38 & 0.08 & 0.39 & 0.16 & 0.39 \\
Edge      & 0.01 & 0.44 & 0.01 & 0.45 & 0.10 & 0.42 & 0.15 & 0.42 \\
SDF (ours)& \textbf{0.30} & \textbf{0.90} & \textbf{0.43} & \textbf{0.91} & \textbf{0.46} & \textbf{0.92} & \textbf{0.53} & \textbf{0.92} \\
\bottomrule
\end{tabular}
\vspace{-0.5em}
\end{table}

\begin{table}[t]
  \centering
  \begingroup
  \scriptsize
  \setlength{\tabcolsep}{4pt}
  \renewcommand{\arraystretch}{0.95}
  \caption{Sensitivity to score-weight choices in Eq.~(\ref{tau}) for $\tau_{\text{obj}}$ selection ($K{=}10$).}
  \label{tab:ablation-tau}
  \resizebox{\columnwidth}{!}{%
  \begin{tabular}{lcccccc}
    \toprule
    Strategy & $(w_c, w_b, w_u, w_{\text{f}})$
             & {\makecell{\textbf{COCO}\\\textbf{AP(50:95)}}} & {\textbf{AP50}} & {\textbf{AP75}} & {\textbf{mIoU}}\\
    \midrule
    Balanced       & $(1.0, 0.6, 0.3, 0.8)$ & 0.53 & 0.80 & 0.66 & 0.92 \\
    Boundary-heavy & $(0.5, 1.0, 0.2, 1.0)$ & 0.53 & 0.78 & 0.65 & 0.92 \\
    Count-heavy    & $(1.5, 0.4, 0.2, 0.6)$ & 0.53 & 0.79 & 0.66 & 0.92 \\
    UIoU-heavy     & $(0.5, 0.4, 1.0, 0.6)$ & 0.53 & 0.79 & 0.66 & 0.92 \\
    \bottomrule
  \end{tabular}%
  }
  \endgroup
\end{table}

\paragraph{Object vs.\ part flexibility.}
Changing only the reference instance masks switches between object- and part-level instances without modifying the backbone or training protocol, a practical advantage when target granularity changes in industrial workflows (Fig.~\ref{fig:object-part}).

\paragraph{Summary.}
Taken together, our experiments demonstrate that BbM is \emph{effective, data-efficient, and flexible} for FSIS of industrial parts, especially in the few-shot regime with very few samples.
First, BbM surpasses strong appearance-driven baselines on $\mathrm{AP}_{50}$/$\mathrm{AP}_{75}$/mIoU under matched splits, with the largest gains precisely where interior cues are weak (texture-poor or color-uniform parts), thereby validating the core premise that \emph{boundaries are the right inductive bias}. 
Second, the accuracy–$K$ curve rises sharply and saturates by $K\!\approx\!2$–$5$, showing that shallow MLP heads already capture boundary structure and that BbM achieves high accuracy with few references—meeting the data-efficiency goal. 
Third, ablations verify that SDF-based boundary supervision is a key driver of BbM, outperforming direct mask/edge targets under identical frozen features and head capacity, while the automatic $\tau_{\text{obj}}$ selection is stable across weight strategies. 
Finally, BbM enables rapid task-level redefinition by swapping only the reference masks, requiring only a short offline preparation ($\sim$2\,min at $K{=}10$).

\section{Conclusion}

We addressed few-shot instance segmentation in settings where the notion of an “instance’’ must be defined by the user and appearance cues are weak. 
We proposed \textbf{Boundary-by-Mask}, which conditions on reference instance masks and learns a boundary-aligned representation via SDF supervision. 
This mask-conditioned, boundary-aware design enables flexible instance selection and task-level redefinition without changing the backbone.

Empirically, boundary supervision proved effective on texture-poor industrial/food objects. A per-pixel shallow MLP head was sufficient to realize the SDF supervision, and performance became stably high once more than about five reference pairs were provided. Additional analyses support the practical viability of BbM. SDF-based boundary supervision is effective, adaptation remains lightweight, and the method is robust to key parameter settings.

\paragraph{Limitations and Future Work.} 
Our current evaluation emphasizes simplified backgrounds; thus, the degree to which the approach transfers to real production environments (clutter, severe occlusion, illumination changes) remains to be verified. Future work includes evaluation on in-the-wild factory data and broader robustness studies, uncertainty-aware reference selection to further reduce annotation effort, leveraging temporal or multi-view cues for boundary stabilization, and integrating boundary supervision more tightly into end-to-end backbones.

{
  \small
  \bibliographystyle{IEEEtran}
  \bibliography{main}
}

\end{document}